\documentclass[10pt,twocolumn,letterpaper]{article}

\usepackage{3dv}
\usepackage{times}
\usepackage{epsfig}
\usepackage{graphicx}
\usepackage{amsmath}
\usepackage{amssymb}
\usepackage{subcaption}
\usepackage[dvipsnames]{xcolor}
\usepackage{booktabs}

\newcommand{\old}[1]{}

\usepackage[pagebackref=true,breaklinks=true,letterpaper=true,colorlinks,bookmarks=false]{hyperref}

\threedvfinalcopy % *** Uncomment this line for the final submission

\def\threedvPaperID{226} % *** Enter the 3DV Paper ID here

\ifthreedvfinal\fi
\begin{document}

%%%%%%%%% TITLE
\title{Towards 3D VR-Sketch to 3D Shape Retrieval}

\author{Ling Luo$^{1,2}$
\and 
Yulia Gryaditskaya$^{1,2}$
\and 
Yongxin Yang$^{1,2}$
\and
Tao Xiang$^{1,2}$
\and
Yi-Zhe Song$^{1,2}$
\and\\
$^{1}$SketchX, CVSSP, University of Surrey $^{2}$iFlyTek-Surrey Joint Research Centre on Artificial Intelligence\\
}

\maketitle
% \thispagestyle{empty}

%%%%%%%%% ABSTRACT
\begin{abstract}
Growing free online 3D shapes collections dictated research on 3D retrieval. Active debate has however been had on (i) what the best input modality is to trigger retrieval, and (ii) the ultimate usage scenario for such retrieval. 
In this paper, we offer a different perspective towards answering these questions -- we study the use of 3D sketches as an input modality and advocate a VR-scenario where retrieval is conducted.
Thus, the ultimate vision is that users can freely retrieve a 3D model by air-doodling in a VR environment.
As a first stab at this new 3D VR-sketch to 3D shape retrieval problem, we make four contributions. 
First, we code a VR utility to collect 3D VR-sketches and conduct retrieval.
Second, we collect the first set of $167$ 3D VR-sketches on two shape categories from ModelNet.
Third, we propose a novel approach to generate a synthetic dataset of human-like 3D sketches of different abstract levels to train deep networks.
At last, we compare the common multi-view and volumetric approaches: 
We show that, in contrast to 3D shape to 3D shape retrieval, volumetric point-based approaches exhibit superior performance on 3D sketch to 3D shape retrieval due to the sparse and abstract nature of 3D VR-sketches.
We believe these contributions will collectively serve as enablers for future attempts at this problem.
The VR interface, code and datasets are available at \url{https://tinyurl.com/3DSketch3DV}.
 
%We propose a regularization loss to explicitly account for abstractness of human 3D sketches. We present a first, to the best of our knowledge, approach to generate a synthetic dataset of human-like 3D sketches. To evaluate our method and the quality of the synthesized data, we collected a test set of $167$ 3D sketches on two shape categories from the ModelNet dataset, using our custom built interface. 
%Encouraged by a growing popularity of sketching in virtual reality settings, in this work we develop a deep model to enable a rough abstract 3D sketch-based retrieval. 
\end{abstract}

%%%%%%%%% BODY TEXT
\vspace{-8pt}
\section{Introduction}
3D model retrieval has become an important topic due to a growing number of free online 3D repositories. It finds applications in CAD design, 3D printing, 3D animation and movies production, where the time required to model a 3D shape can be strongly reduced by relying on retrieval from existing 3D shape collections. Various input modalities have been tried -- images \cite{lee2018cross} and rough 2D sketch \cite{wang2015sketch} at first, with latest research focusing on 3D to 3D, i.e., using 3D scans \cite{avetisyan2019end, dahnert2019joint} or existing 3D shapes \cite{su2015multi, qi2016volumetric, ma2017boosting, feng2018gvcnn,  grabner20183d, grabner2019location, he2019view,qi2017pointnet, qi2017pointnet++, uy2020deformation}. 

Despite great strides made, two salient questions still remain (i) what constitutes the best input modality to initiate 3D retrieval, and (ii) under what usage scenario can such retrieval be best facilitated. 
For the former, 2D sketches and images are both in 2D, therefore can not reach the level of details desired, and the 2D-3D domain gap can also be counter-intuitive. The 3D-based paradigm on the other hand dictates existing 3D models to be readily available, which to some degree forms a ``chicken-and-egg'' problem (i.e., where/how to source the input model at the first place). As for the latter, apart from 2D sketches which are freely defined by the user, all other usage scenarios do not offer flexibility in terms of the input desired -- one can not easily alter an image, not to mention a 3D model/scan.

In this paper, we offer a new perspective on 3D shape retrieval -- we advocate the use of 3D sketches as a new input modality.
This new 3D-VR sketch to 3D shape paradigm not only enables detailed retrieval as per the common 3D model to 3D model setting, but also simultaneously offers a degree of flexibility found elsewhere in 2D sketch-based retrieval. Our ultimate vision is as follows: with a specific 3D model in mind, one emerges into a VR environment, roughly sketch out the model using handheld controllers, press retrieve and then relevant models would start populating the VR environment -- for any downstream tasks the user may desire (e.g., VR shopping, 3D printing).

\begin{figure}[t]
    \centering
    \vspace{-4pt}
    \includegraphics[width=\linewidth]{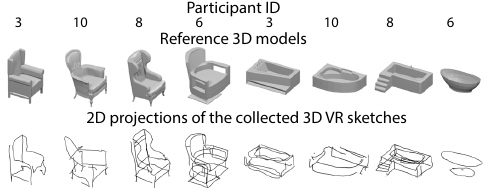}
    \caption{Our collected VR sketches exhibit variability of sketching styles and levels of details.}
    \label{fig:vr_sketches}
    \vspace{-12pt}
\end{figure}

Our first contribution is therefore coding a VR environment for the said purpose. With this VR environment, we collected the first human VR-sketch dataset. $10$ users with no artistic background were hired to produce a total of $167$ 3D VR-sketches from the two categories from ModelNet: chairs and bathtubs. Some examples are shown in Figure \ref{fig:vr_sketches}.
Since the collection process is both time- and cost-sensitive, we additionally propose the first 3D model to 3D sketch generator, and construct a synthetic dataset of 3D sketches. 
A key trait of our generator is that it can produce 3D sketches of different abstraction levels, effectively simulating that found in real-human sketches. Through training a series of deep 3D VR-sketch to 3D retrievals model using the real and synthetic datasets, we drive out a few important insights (i) retrieval performance drops with an increasing abstraction level, and (ii) models trained with synthetic sketches can already reach a decent performance level when tested on human sketches. 

At last, we experiment with different shape/sketch representations, and state-of-the-art losses commonly used in 3D model retrieval~\cite{he2018triplet,he2019view,uy2020deformation} but re-purposed for our problem. We focus on investigating the domain gap between 3D shapes and 3D sketches (see Figure~\ref{fig:vr_sketches} for an example), due to the two key properties of sketches: \emph{sparsity}: full 3D models versus sparse sketched lines, and \emph{abstraction}: 3D models are geometrically perfect, while sketches are subject to deformations. 
In particular, we examine multi-view versus point-based 3D representations, and show that the latter is more robust to both sparsity and an increased level of abstractness. 
We further propose an architecture with a reconstruction path to explicitly allow for change in abstractness. 

%We introduce a reconstruction track, that can be seen as a regularization term for the input sketch, where the source is an input sketch and the target is a clean network of strokes. 

In summary, our contributions include: (i) a new perspective on 3D model retrieval, where 3D VR-sketches are used for the first time to conduct retrieval, (ii) a dataset of human 3D VR-sketches, collected using a purpose-built VR environment, (iii) an approach to generate a synthetic 3D sketch with a variable level of abstractness, (iv) comprehensive evaluations using recent 3D shape retrieval models re-purposed to the task of 3D sketch-based retrieval, to drive out insights, plus a novel regularization track to address the sketch-model domain gap.

\section{Related work}
%\subsection{3D Sketch in VR}
%Arora et al.\cite{arora2017experimental} compare traditional sketch on physical surface and sketching in VR and evaluate how much visual guidance can help compensate the inaccuracies of VR sketching.
%Giunchi et al. \cite{giunchi2019mixing} also compared 3D/2D sketch interaction and show that 3D mid-air sketching is more suitable experience within VR even overcoming the familiarity with 2D based retrieval.
%Rosales et al.\cite{rosales2019surfacebrush} propose a novel 3d modeling approach by drawing ribbon-like 3D brush stroke in VR.
%\yulia{I think we do not need this section, instead the relevant work should be mentioned in the introduction.}

\paragraph{3D shape representation.} 
When dealing with 3D shape retrieval from a single image or a 3D shape, existing work is divided into two large groups based on the shape representation used: view-based \cite{su2015multi, qi2016volumetric, ma2017boosting, feng2018gvcnn,  grabner20183d, grabner2019location, he2019view} or volumetric. The volumetric representation can be further broken down to point-cloud \cite{ravanbakhsh2016deep, qi2017pointnet, qi2017pointnet++, uy2020deformation} or voxel-based \cite{wu20153d, maturana2015voxnet, brock2016generative, lin2018learning}. 
In this work we show that on the task of 3D sketch to 3D model retrieval point-cloud representation beats multi-view approaches due to better handling of the sparsity of 3D sketches.

\paragraph{2D image- and 3D shape-based retrieval.}
The retrieval problem in a multi-class setting is closely related to the shape classification problem, where the intermediate embedding of an image or a 3D shape is used for retrieval.
A vanilla approach for multi-class shape classification is to use a softmax cross-entropy loss~\cite{he2019view,liu2018cross}. Others works specifically tackle retrieval \cite{su2015multi,he2018triplet}, where 
%the goal is to capture well both inner-class and inter-class differences.  
%For example, the Mahalanobis distance was adopted in .
the triplet loss and its variants \cite{wang2014learning, yu2016sketch, wen2016discriminative,he2018triplet} have become a common standard.
%Typically, a sample from the same class is used as a positive sample, and a sample from a different class or a distant sample form the same class is used as a negative sample \cite{wang2014learning, yu2016sketch}. 
%The disadvantage of the triplet loss is a costly procedure of triplet mining.
%Schroff et al.~\cite{ schroff2015facenet} substituted a hard triplet mining with a semi-hard triplet sampling strategy to avoid local minima early on.  
% Alternatives have been proposed to address the triplet sampling problem. \cite{wen2016discriminative} introduced a Center Loss that learns the center of each class in a feature space and minimizes the distances between the class samples features and the respective center of the class. 
%The drawback of this loss is that there is no condition that insures that the classes do not  overlap between each other. 
Amongst these, 
\cite{he2018triplet} represents the state-of-the-art for 3D model retrieval. It combines triplet and center losses \cite{wen2016discriminative}, to solve for their respective drawbacks by using the class center as a positive sample. 
%Instead of mining for hard/semi-hard negative triplets they take as a negative the center of the closest class in a feature space.
In this work we evaluate both the triplet and the triplet center losses on the task of 3D sketch-based retrieval and combine them with an additional Reconstruction Loss, tailored to the 3D sketch-based retrieval problem.

Due to a domain gap between images and 3D shapes, or target and query 3D models being from different distributions, Siamese \cite{yu2016sketch, qi2016sketch} or Heterogeneous \cite{ye20163d,lee2018cross,liu2018cross} network architectures can be more beneficial for a certain problem.
%
% Another strand of research focus on learning a joint embedding to conduct retrieval. They can be broadly categorized into Siamese \cite{yu2016sketch, qi2016sketch} or Heterogeneous \cite{ye20163d,lee2018cross,liu2018cross}.
% % Salient examples for image to 3D shape retrieval include Lee et al.~\cite{lee2018cross}, who adopt a triplet loss, and propose a fully-connected adaptation layer after the image CNN branch. 
% %They additionally proposed a more efficient triplet sampling strategy, by constructing triplets after the network forward pass.
% \cite{hong2016multi,liu2018cross} reduces the domain shift by projecting the initial encoding obtained by training classification loss on a single domain into new subspaces.
%
In this work, we compare these two types of architectures for our multi-view baseline, and show that due to a strong domain gap between a 3D sketch and a 3D shape, the Heterogeneous architecture by far beats the Siamese one.

% Recently Jing et al~\cite{jing2020self} proposed a self-supervised method by training their architecture to classify if the features extracted using multi-view representation based on ResNet18 \cite{he2016deep} and point cloud representation \cite{wang2019dynamic} are from the same object or not. In addition, they use the triple loss to obtain view-invariant features for a multi-view shape representation.

\paragraph{3D sketch-based 3D model retrieval.}
So far there was very little work on 3D sketch-based retrieval \cite{li20153d, li2016shrec, ye20163d, giunchi20183d}, especially in recent years. Most of them work with sketches collected using Microsoft Kinect, which not only has limited tracking accuracy, but the collection interface is also counter-intuitive, where 3D sketching is performed while visualizing 2D projections. As a result, sketches collected mostly have low fidelity and exhibit less details. Our VR-sketches are completely different: (i) visualization and sketching are both performed in 3D, and (ii) the use of the latest VR technology offers a high precision. Together, they ensure our VR-sketches are of high-fidelity and rich in details, thus more fitting for retrieval. 
A notable exception is the work by Giunchi et al.~\cite{giunchi20183d}, yet they address a different problem of 3D model retrieval for dense-color VR sketches (and optionally base 3D shapes), while we target at a more simplistic and abstract shape representation -- a sparse set of single-color lines.
%they our VR-sketche are much more Motivated by a recent intersect in 3D sketching in a VR environment, we collect a first dataset of 3D sketches, using a VR headset and a custom-build interface that enables detailed saving of space-time sketch coordinates, for two shape categories. 
Note that we can not find the said dataset \cite{li20153d, li2016shrec, ye20163d} in an open access, thus can not offer a direct comparison. Our NGVNN \cite{he2019view} baseline is however already superior to the state-of-the-art used by \cite{ye20163d} and \cite{giunchi20183d}.

\paragraph{Non-photorealistic rendering (NPR).}
NPR is an old graphics and vision problem, for a detailed overview of existing methods for generating 2D NPR rendering we refer the interested reader to a recent report \cite{B_nard_2019}. 
% NPR rendering is often used to generate synthetic training for 2D sketch-based retrieval and reconstruction tasks \cite{yoon2010sketch,shao2011discriminative,li2013sketch_a,li2013sketch_b,gryaditskaya2019opensketch}.
The only attempt to produce the representation that resembles a 3D sketch automatically form a 3D shape was proposed by Li et al.~\cite{li20153d} as a concatenation of the shape views from six canonical viewpoints. They use this representation to show that the performance of their non-learning method on outline to 3D model retrieval task significantly outperforms the performance of the 3D sketch to 3D model retrieval. This experiment indicates that such simple shape representation is not sufficient as training data for 3D sketch-based 3D shape retrieval. Our experiments support that: The network trained on sketches with higher abstractness values significantly outperforms the network trained on clean sketches.
To the best of our knowledge, we are the first to propose a method to generate abstract 3D sketches from 3D models.

%\comment{The related work needs to be shortened, once we have a full picture.}

\section{Datasets}
Collecting a full dataset of 3D human sketches is a labor intensive task. 
We collect a small dataset of human sketches that we use to guide the synthetic dataset generation. We as well use it as a test set to validate that the network trained on the proposed synthetic data generalizes well to human sketches.
To obtain the training data, we adopt a common strategy of generating a synthetic training data.

\subsection{Selected shapes}
% * Describe datasets\\
% * Describe selected classes \\ 
For training and testing our sketch-based retrieval models we use the repaired clean manifold meshes\footnote{\url{https://github.com/lei65537/Visual_Driven_Mesh_Repair}} \cite{Chu:2019:MeshRepair} from ModelNet10. % and ShapeNetCore55.
We use all 10 classes from ModelNet10, with the total of 3958 shapes. We split the dataset into train, validation and test sets, which contain 2847, 317 and 792 shapes, respectively, ensuring a proportional split of shapes of each class between three sets.

%\textbf{ShapeNet40}: We select 40 classes from the 57 classes of ShapeNet, excluding those that have too few  shape samples, with the total of 3369 shapes.

%yulia: instead of curved networks use shapes, since at that moment in the paper there were nothing about the curve netwroks yet.

%We use synthetic sketch-shape pair to train and test all our baselines.
%yulia: That is not the place, should be closer to experiments.
\subsection{Human 3D sketch dataset}\label{sec:human_sketch}

\paragraph{Task.}
We are targeting 3D sketches created by novices, which can be viewed as an equivalent of the quick 2D sketches from the QuickDraw dataset~\cite{ha2017neural}. To enable the usage of the collected sketches for fine-grained retrieval evaluation we built a dataset of paired sketches and 3D models.
We experimented with a setting, where one can observe a 3D model in VR for an unlimited duration of time, and then is asked to sketch from memory. 
We observed that under this scenario the participants were sometimes omitting features important to accurately testing fine-grained retrieval, and instead let the participants to sketch over a reference 3D model (Figure \ref{fig:vr_sketch}).
We provide in the supplemental a qualitative evaluation of the sketches from memory and the retrieval performance on such inputs.
To mimic the level of detail that can be expected from the 3D sketches from the imagination, we opted to use wide ribbon lines, but do not pose any constraints on sketching style or level of detail.

\paragraph{Participants.}
We selected 139 chairs and 28 bathtub shapes from ModelNet10's test set, and hired 10 participants, who have no art background or 3D sketch experience. The shapes were randomly divided into 10 subsets, consisting of 13-14 chairs and 2-3 bathtubs each. 
Each participant was assigned with one of these subsets. There is no duplicated shapes between subsets.

\begin{figure}[t]
    \centering
    \vspace{-4pt}
    \includegraphics[width=\linewidth]{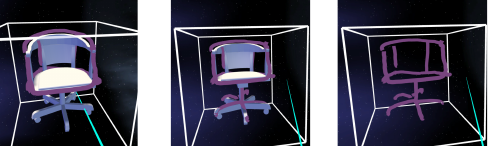}
    \caption{Our VR sketching environment allows to load the reference 3D model, and the user is asked to sketch around it. At any moment the user can hide the 3D model.}
    \label{fig:vr_sketch}
    \vspace{-12pt}
\end{figure}

\begin{figure*}[t]
    \centering  
    \vspace{-4pt}
    %\fbox{\rule{0pt}{2.0in} \rule{\textwidth}{0pt}}
    \includegraphics[width=\linewidth]{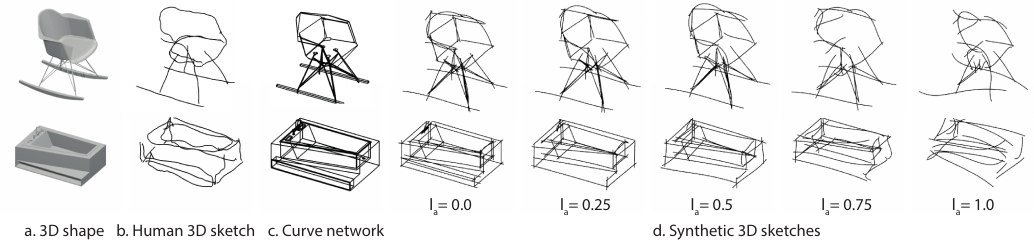}
    \caption{For a subset of shapes from the two categories from ModelNet10: chairs and bathtubs (a), we collect human novices sketches (b). We generate a synthetic dataset of 3D sketches for the 10 shape categories from ModelNet10. We first generate detailed curve networks with FlowRep~\cite{gori2017flowrep} (c) and then apply a set of detail and stroke filtering steps to mimic the appearance of human sketches, where the appearance is controlled by an abstractness parameter $l_a$ (d), defined in Section \ref{sec:abstraction}.}
    \label{fig:datasets}
    \vspace{-10pt}
\end{figure*}

\paragraph{Interface.}
%Since none of the existing VR design software tools, such as TiltBrush or Quill, allows to load a reference shape or support detailed saving of the sketching process, we implemented our custom 3D sketching environment based on Oculus Rift platform and Unity engine. We save time-space coordinates of each stroke.
Although there are general-purpose VR painting and design softwares that enable users to draw directly in 3D (such as Google's Tilt brush\footnote{\url{https://www.tiltbrush.com/}}, and Facebook's Quill\footnote{\url{https://www.quill.com/}}), they do not serve our purpose in full: (i) we would like to record detailed stroke-based information, (ii) we require the option of displaying a reference model for data collection, and (iii) we want a fresh code base for any additional functionalities in the future (e.g., sketch-based 3D editing). We implemented our custom 3D sketching environment based on the Oculus Rift platform and Unity engine.

\subsection{Synthetic training data generation}
\label{sec:data_generation}
%We first convert all the shapes to watertight meshes using the method of Huang et al.~\cite{huang2020manifoldplus}. 
As a first step towards obtaining a synthetic sketch representation of a 3D shape, we extract detailed curve-networks with the method of Gori et al.~\cite{gori2017flowrep}. This method is designed to produce a curve network that preserves well shape details, which are, though, uncommon for human novices sketches (Figure \ref{fig:datasets} b,c). 
We observe that novices not only omit small details, but tend to represent thin volumetric details with single lines and moreover often omit subset of feature lines (Figure \ref{fig:datasets} b).
To obtain human sketch appearance, we focus on two aspects: level of detail and mechanical inaccuracies. 
We first perform details filtering and lines consolidation, followed by lines filtering. We then break the long curve chains into shorter strokes, to which we apply a set of local and global transformations (Figure \ref{fig:datasets} d).

\subsubsection{Curves network extraction.}
%* Describe the coverage parameter and its settings \\
The FlowRep method~\cite{gori2017flowrep} requires an input to be a curvature-aligned quad-dominant mesh. For processing efficiency, we simplify original triangular meshes by decimation before converting them into quad-dominant meshes using Blender, but the quality of conversion does not always comply with the requirement of FlowRep. Thus, in this stage, we filtered nearly 20\% of original dataset which failed to be processed by FlowRep. This can be alleviated in the future by relying on more advanced quad-meshing algorithms \cite{lyon2019parametrization}.

\begin{figure}[ht]
    \vspace{-10pt}
    \centering
    \includegraphics[width=\linewidth]{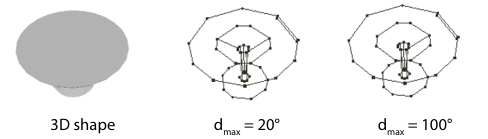}
    \caption{We found little effect of the descriptiveness threshold parameter $d_{max}$ in FlowRep~\cite{gori2017flowrep} on most of the shapes from ModelNet10 dataset.}
    \label{fig:FlowRep}
    \vspace{-14pt}
\end{figure}

FlowRep allows to control the level of detail of the output curve network by adjusting the values of the descriptiveness threshold $d_{max}$. We, nevertheless, found little impact of this parameter on the output networks, as illustrated in Figure \ref{fig:FlowRep}.
We set this parameter to $20^{\circ}$ to generate our synthetic dataset, following their default value. 

%The larger values lead to less detailed networks, nevertheless 

%The larger values lead to less detailed networks, as illustrated in Figure \ref{fig:FlowRep}. 
%\clarify{To test the affect of sketch diversity, we also generate a version with coverage equals to 50}
%\yulia{How do we use currently sketches with this parameter set to 50?} \todo{Maybe add the figure that shows how this parameter affects the result? We also need to explain why do not we use a large value and instead do all the processing steps}
%As shown in Fig. \ref{fig:corverage}, the coverage parameter has minor impact on abstraction, most details are still kept. So we designed details filtering and consolidation steps to simplify the networks.
%

\subsubsection{Details filtering and consolidation.}
\label{sec:network_simplify}
The code by Gori et al.~\cite{gori2017flowrep} returns networks of curves in a form of chained edges of an input mesh. We first break these chains into several if the angle between two consequent edges is smaller than 135 degrees. We then remove all the chains whose length is smaller than $10\%$ of the smallest of height and width of the bounding box of the original shape ($d_{min}$). We re-sample all the chains using Ramer-Douglas-Peucker algorithm \cite{ramer1972ramer} with an accuracy parameter set to $0.02 \, d_{min}$. Finally, we iteratively go over all the chains and compute for each chain the closest, tangentially aligned chain. If the distance between such two chains is smaller than $5\%$ of the largest length of the two chains, the chains are substituted by their aggregate curve. These steps allow to remove small details and lines that are close to each other. Yet, the simplified curve networks contain many more lines than can be found in majority of 3D sketches by novices. 

\subsubsection{Abstraction.}\label{sec:abstraction}
In this section we describe our approach to obtain a 3D sketch with different levels of detail and mechanical inaccuracies, which we jointly refer to as a level of sketch abstractness $l_a$. We constraint $l_a$ to be between 0 and 1.

\paragraph{Level of detail}
(I) To reduce the number of lines we first compute the similarity matrix using a discrete Frechet distance among chains. When computing the Frechet distance we first align the strokes at one of their end points, by translating one of the strokes.
We then perform the grouping according to this similarity matrix with Agglomerative Hierarchical Clustering, where the number of cluster $n_{cluster}$ is a function of $l_a$ and the number of chains in the sketch $n_{chains}$: $n_{cluster} = \max(n_{chains}(1-0.8 l_a)/2, 10)$.
(II) We next for each cluster iteratively select a pair of two most distant lines in terms of their absolute positions and compute the mean distance from all the lines $d_{mean}$ in the cluster to these two lines. We remove all the lines in a cluster for which the distance to any of the two selected lines is less than $d_{mean}$.
%To this end, we, first, set a parameter $l_a \in [0, 1]$ to control the abstraction degree. For synthesis, we choose 5 values for this parameter to generate 5 levels of abstraction, $[0.0, 0.25, 0.5, 0.75, 1.0]$, and for each value, we remove redundant edges correspondingly. That is to say, we keep 100\%, 80\%, 60\%, 40\%, 20\% of original edges respectively for these 5 levels. To achieve this, we first compute the similarity matrix using discrete Frechet distance and distance matrix among chains. Then we group these chains by similarity matrix with Agglomerative Clustering. The number of clusters $n_{cluster}$ is decided by $l_a$, $n_{cluster} = max((1-0.8\times l_a)/2, 10)$.
% \clarify{Next, we sample no more than 4 chains from each cluster by selecting remotest chain pairs and remove chains that are too close to the selected ones repeatedly.}
(III) After getting the reduced set of chains, we break long chains into several shorter chains, which we refer to as strokes. We split each chain at vertices, where the curvatures are twice larger than the mean curvature of an original chain. We further filter out short strokes whose lengths are smaller than $0.2s_{max}$ to avoid tiny details, where $s_{max}$ is the largest dimension of an input network.

%\begin{figure*}[ht]
%    \centering
%    %\includegraphics[scale=0.4]{image/siamese_2branch.png}
%    \includegraphics[width=0.9\textwidth]{image/networks.pdf}
%    %\caption{Network architectures for view-based baselines including 2D sketch to 3D shape and 3D sketch to 3D shape. Image Encoder also use VGG-11 but change the output embedding length to 512 to keep consistent with NGVNN-based 3D shape branch. }
%    \caption{The diagrams on the left demonstrate the siamese and heterogeneous network architectures with NGVNN as a backbone. The diagrams on the right show the networks based on PointNet++, when trained as described in Section \ref{sec:metric_learning} (top), and in Section \ref{sec:regularization_branch} (bottom).}
%    \label{fig:networks}
%    \vspace{-10pt}
%\end{figure*}

\begin{figure}[ht]
    \centering
    \includegraphics[width=1.0\linewidth]{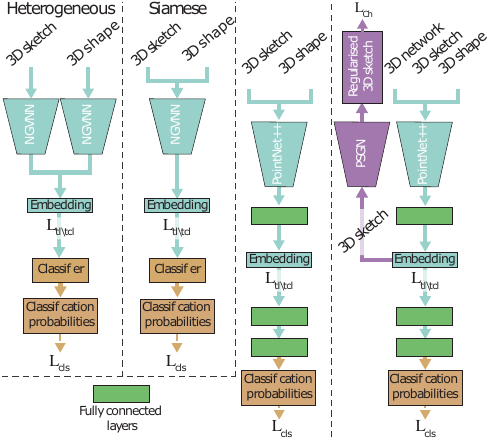}
    %\caption{Network architectures for view-based baselines including 2D sketch to 3D shape and 3D sketch to 3D shape. Image Encoder also use VGG-11 but change the output embedding length to 512 to keep consistent with NGVNN-based 3D shape branch. }
    \caption{The two diagrams on the left demonstrate the heterogeneous and siamese network architectures with NGVNN as a backbone. The two diagrams on the right show the networks based on PointNet++, when trained as described in Section \ref{sec:metric_learning} and in Section \ref{sec:regularization_branch}.}
    \label{fig:networks}
    \vspace{-10pt}
\end{figure}

\paragraph{Mechanical inaccuracies} 
To mimic mechanical and perspective inaccuracies of human sketches we apply to each stoke a set of global and local deformations. 
We first apply a global translation, rotation and scaling, depending on the given level of abstraction $l_a$. To achieve this we deploy an auxiliary parameter $t$, which is randomly sampled from the range $[0, 1.5 l_a]$. The rotation angle for each axes is then independently randomly sampled from the interval $[-10 t, 10 t]$ degrees. The scale factor for each axes is independently randomly sampled from $[1 - 0.1 t, 1 + 0.1 t]$.
To obtain a translation vector we randomly sample from the surface of a sphere with its radius value randomly sampled from $[0, s_{max} t]$, where $s_{max}$ is the largest dimension of an input network, as before.
% Global transfromation for stroke is composed of 5 steps:
% \begin{enumerate}
%     \item Translate to zero center
%     \item Rotation: rotation angle for each axis is sampled from $[-10\times t, 10\times t]$ degree respectively,.
%     \item Scale: scale factor for each axis is sampled from $[1-, 1+0.1\times t]$ respectively.
%     \item Translate back to original center
%     \item Translate: we random sample a translation vector from a global surface whose radius $r$ samples from$[0, S \times t]$ where $S$ is the max scale of input network. %For x-y plane, first choose a random angle $\theta$ from$[-180, 180]$degree, then sample vector length$d$ from $[0, S_{max} \times t]$, then translation for x $t_x=d\times \cos \theta$, translation for y $t_y=d\times \sin \theta$, where $S_{max}$ is the max scale of input network. For z axis, we sample $t_z$ from $[0, S_{max} \times t]$.
% \end{enumerate}
After a global stroke deformation, we apply a random translation for each stroke vertex. The translation vector is randomly sampled from the disk with a radius $r_{v_i}$, which lies in the plane orthogonal to the stroke direction in the vertex ${v_i}$. The radius $r_{v_i}$ is randomly sampled from the range  $[0, 0.1 l_a l_{stroke}]$, where $l_{stroke} \in [0,1]$ is the length of the current stroke.
%\clarify{how is that done, what are the parameters?}
%We map global stroke transformation and local vertices displacements to a single control parameter $l_a \in [0,1]$, that represents the level of abstraction.
%$R_{trans} = 1.5 l_a$ and $R_{noise} = 0.1 l_a $.
%\clarify{How the mapping is implemented?} 
%Setting this parameter $l_a$ to $0$ corresponds to a filtered curve network.  
We extend both ends of each stroke by $p$ which value is randomly sampled from $[0, 0.1s_{max}]$, to reproduce human strokes appearance.

After global and local stroke deformations, we apply 3D spline interpolation, which results in smoother strokes, matching the appearance of  human strokes.
\\

For our experiments, we generate 5 synthetic datasets with 5 levels of abstraction, $[0.0, 0.25, 0.5, 0.75, 1.0]$, and obtain two mixed datasets by mixing sketches of all abstraction levels or only the sketches with 3 abstraction levels in the middle.
Figure \ref{fig:datasets} c.~shows example sketches obtained with different settings of $l_{a}$ parameter.

\section{Networks}

\subsection{Backbones}
\paragraph{Multi-view representation}
We choose View N-Gram Network (NGVNN) \cite{he2019view} as a baseline for a multi-view based shape retrieval, since it was shown to outperform the alternative solutions \cite{li2019angular,he2018triplet,feng2018gvcnn,su2015multi} on the 3D-shape based retrieval task on ModelNet40. We employ VGG-11 with batch normalization \cite{simonyan2014very} as a backbone. The output of the penultimate layer is used as a feature embedding. We experiment with two architectures: one where the same network is used to encode the 3D sketch and 3D shape (Siamese), and one where there are two branches whose weights are trained separately (Heterogeneous) (see Figure \ref{fig:networks}).

\paragraph{Volumetric representation}
Due to sparsity of a 3D sketch, among the volumetric shape representations we opted for point-based.
%* Quick description of the baseline.
In particular, we select PointNet++ \cite{qi2017pointnet++}, which applies hierarchical feature learning to efficiently capture local structures and to get a robust point cloud representation. We adopt their multi-scale grouping and random input dropout during training (MSG+DP) strategy, which is more robust to a variable sampling density on an input point cloud due to an adaptively selected neighborhood's size. We use the same three-level hierarchical network with three fully connected layers as in the original paper and use the output of the first fully connected layer as a feature embedding. We use the Siamese network for this model.
\\

We train both PointNet++ and NGVNN from scratch for a fair comparison. The feature vector is a 512-dimensional vector in both cases.
%For a fair comparison between PointNet++ and NGVNN, we initialize the backbone without pre-training since there's no pretrianed model for PointNet++.
%\clarify{What is the original PointNet++ and in NGVNN?}

\subsection{Losses}
\label{sec:metric_learning}
We compare the performance of the two backbone 3D representations with the optimal sampling and rendering strategies, when trained with the classification loss $\mathcal{L}_{cls}$, versus training with a combinations of the classification loss and the triplet loss $\mathcal{L}_{tl}$ \cite{wang2014learning}, or the combination of the classification loss and the triplet center loss $\mathcal{L}_{tcl}$ \cite{he2018triplet}:
\begin{equation}
    \mathcal{L} = \lambda_{cls}\mathcal{L}_{cls} + \lambda_{tl\backslash tcl}\mathcal{L}_{tl \backslash tcl}\label{math:cls_tl},
\end{equation}
we set $\lambda_{cls} = 1$ and $\lambda_{tl} = \lambda_{tcl} = 0.1$.

The goal of both triplet loss and triplet center loss is to obtain a better structured embedding space. 
%We train each model as a feature extractor followed by two fully connected classification layers. 
The input for the cross-entropy loss $\mathcal{L}_{cls}$ are the class probabilities, while the input for the triplet loss and triplet center loss are features extracted by the encoders (Figure \ref{fig:networks}). 

\paragraph{Triplet loss}
The goal of the triplet loss is to ensure that the anchor-negative distances are larger than the anchor-positive distances by a given margin. For the triplet loss $\mathcal{L}_{tl}$, we use both 3D sketches and 3D shapes as anchors. 3D shapes and 3D sketches sharing the same class as the anchor are treated as positive, while those from different classes are negative samples. The triplet loss for the $i$-th triplet is:

\begin{equation}
    \mathcal{L}_{tl}^i = \max\{0, d_{pos}^i - d_{neg}^i + m_{tl}\},
    \label{math:triplet}
\end{equation}
where $d_{pos}^i$ is the anchor-positive distance and $d_{neg}^i$ is the anchor-negative distance, $m_{tl}$ is the margin. As a distance $d$ we use the squared Euclidean distance, where the feature space is normalized to a unit hypersphere. We choose the margin value by calculating the average difference between $d_{pos}$ and $d_{neg}$ of the validation set of the model trained with the classification loss only. We set margin for NGVNN baseline to $m_{tl}= 2.0$ and for PointNet++ to $m_{tl} =1.8$.
%For NGVNN baseline, we set margin to $m_{tl}= 2.0$, and for PointNet++, we set margin to $m_{tl} =1.8$.

\paragraph{Triplet center loss}
Triplet center loss~\cite{he2018triplet} combines the strength of the triplet loss and the center loss. It avoids the problem of triplet hard mining by taking as a positive the learned center of the shape's class. As a negative sample it takes the closest center of some other class: 
\begin{equation}
    \mathcal{L}_{tcl}^i = \max\{0, d(f^i, c_{y^i}) - \min_{j\neq y^i}d(f^i, c_j) + m_{tcl} \} \label{math:triplet_center},
\end{equation}
where $y^i$ is a class label of the anchor 3D sketch/shape, $f^i$ is the embedding of an anchor and $c_k$ is the center of the class $k$ in the embedded space.%, and $d$ is the squared Euclidean distance.
\\

Given the data imbalance among classes of the used 3D shape dataset, we employ a balanced batch sampler. For each batch, we first randomly select $k=8$ classes from all $K$ classes, then we select $p=1$ 3D sketch-shape pairs for each of the $k$ class. For efficiency, we use an on-the-fly hard negative triplet mining to calculate the triplet loss rather than generating the triplets beforehand. 
%For all our experiments, subject to the GPU capacity, we set $n=8$ and $p=1$. 

%We filter out the triplets whose values are less than zero and only keep the hardest negative for each anchor-positive pair as valid triplet. 
The final loss $\mathcal{L}_{tl/tcl}$ is the mean of all triplet (center) losses within a batch:
%We denote $\mathcal{L}_{ij}$ the for the i-th sketch anchor, j-th positive shape and corresponding hardest negative shape. Each triplet loss is computed by Eq.\ref{math:triplet}.
% $$
% \mathcal{L}_{tl} = 1/N \sum_{i=1}^n \sum_{j=1}^{m} ReLU(l_{ij})
% $$
% \begin{equation}
$\mathcal{L}_{tl \backslash tcl} = \frac{1}{N} \sum_{i=1}^N  \mathcal{L}_{tl \backslash tcl}^i,$
% \end{equation}
where $N$ is the number of all valid triplets.
%, and $\mathcal{L}_{tl}^i$ is a loss of an $i$-th triplet, given by an Equation \ref{math:triplet} .

% Given dataset $\{(x^i, y^i)\}_{i=1}^N$ which contain $N$ samples $x^i$ with corresponding labels $y^i\in\{1,2,\dots,C\}$, $C$ is the number of classes. We use $f_\theta(\cdot)$ to represent the encoder that maps shapes or sketches into $d$-dimensional vectors $f(x_i)$ and use $f^i$ to represent $f(x_i)$.

% While triplet loss $L_{tl}$ maximize the difference between anchor-positive pairs and anchor-negative pairs, triplet center loss $L_{tcl}$ maximize the difference between anchor-positive center pair and anchor- negative center pair:

\subsection{Regularization branch} 
\label{sec:regularization_branch}
To account for a variation of human styles, we aim at pushing the 3D sketch embedding towards style invariance by introducing a regularization branch. We propose the architecture with an additional decoder branch (Figure \ref{fig:networks}), based on the reconstruction point-based architecture (PSGN) by Fan et al.~\cite{fan2017point}, and only use 2 fully connected layers to form the reconstruction branch. As a reconstruction loss, we use the Chamfer distance $\mathcal{L}_{Ch}$ between an input 3D sketch with a given level of abstraction and the 3D curve network obtained as described in Section \ref{sec:network_simplify}.
The full loss is
\begin{equation}
    \mathcal{L} = \lambda_{cls}\mathcal{L}_{cls} + \lambda_{tl\backslash tcl}\mathcal{L}_{tl \backslash tcl} + \lambda_{Ch}\mathcal{L}_{Ch},
\end{equation}
where $\lambda_{cls} = 1$ and $\lambda_{tl\backslash tcl} = 0.1$, as before. We choose $\lambda_{Ch}=8$ for TCL and $\lambda_{Ch}=12$ for TL.

\begin{figure}[htb]
    \vspace{-10pt}
    \centering  
    % \fbox{\rule{0pt}{1.0in} \rule{0.9\linewidth}{0pt}}
    \includegraphics[width=\linewidth]{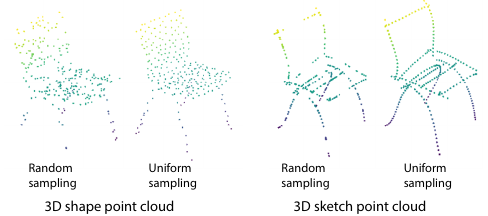}
    \caption{Visual comparison of random and uniform sampling strategies. To clearly show the difference, we set the number of points to 256 for this visualization.}
    \label{fig:sampling}
    \vspace{-10pt}
\end{figure}

\begin{figure*}[t]
    \centering
    \includegraphics[width=\linewidth]{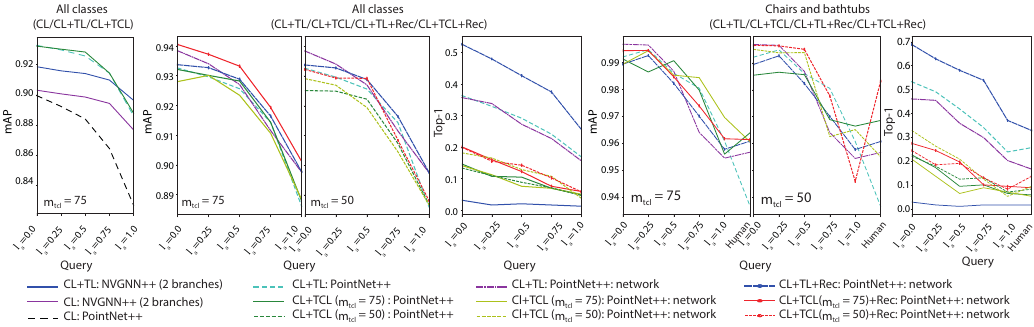}
    \caption{mAP values and Top-1 scores using models trained with the dataset of 3D sketches with  $l_a = 0.5$. When we train PointNet++ with $m_{tcl} = 75$, the value that optimizes the validations synthetic dataset, then CL+TCL+Rec gives the highest mAP over all the baselines on all synthetic datasets, but has slightly lower mAP on human dataset than the network trained with CL+TCL. When $m_{tcl} = 50$, the value that optimizes the mAP on a human dataset of the network trained with CL+TCL, the reconstruction loss (Rec) boosts the mAP on human dataset from $0.969$ to $0.983$. ':network' means that the network, obtained as described in Section \ref{sec:network_simplify}, is contributing to the triplets construction.} 
    \label{fig:baselines}
    \vspace{-8pt}
\end{figure*}

\section{Evaluations}
We adopt the following common evaluation measures to evaluate the retrieval performance: Mean Average Precision (mAP), Normalized Discounted Cumulative Gain (NDCG), Nearest Neighbor (NN), which evaluate the ability of a model to discriminate between shape classes, and Top-k accuracy, which measure how many of the retrieval tasks have a ground-truth within top k retrieved results.

\subsection{Sampling and rendering strategies}
\label{sec:sampling_rendering}
We experiment with two rendering styles for  NGVNN \cite{he2019view} and two sampling strategies for PointNet++ \cite{qi2017pointnet++}.

\paragraph{3D sketch rendering for multi-view network.}
We generate 12 $224\times224$ orthographic views of each 3D shape and 3D sketch by placing 12 virtual cameras around it every 30 degrees, as was proposed by Lee et al.~\cite{lee2018cross}. The cameras are elevated 30 degrees from the ground plane. For both 3D shapes and 3D sketches we experiment with two types of rendering styles: Phong Shading and depth maps.
%(Figure \ref{fig:rendering_views}). 
For 3D sketches we represent each line as a 3D tube. 

\paragraph{Point cloud sampling.}
To get the point cloud representation, we first sample 10000 points from shapes and 3D sketches. For shapes we use Monte-Carlo sampling\footnote{\url{https://github.com/fwilliams/point-cloud-utils}} and for sketches we use equidistant sampling. We then adopt two types of sampling from the initial 10000 points to obtain a sparse set of 1024 points: random sampling or uniform sampling. The uniform sampling is obtained with the farthest point sampling. The sparse sets are obtained on-the-fly and thus might differ from one iteration to another. Figure \ref{fig:sampling} shows the visualization of random and uniform sampling.

% \subsection{Evaluation measures}
% In our experiments, we adopt the following common evaluation measures to evaluate the retrieval performance: Mean Average Precision (mAP), Normalized Discounted Cumulative Gain (NDCG), Nearest Neighbor (NN) and Top-k accuracy.

% \begin{description}
%     \item[Mean Average Precision (mAP)] is the mean of the average precision scores for each query, where the average precision is the area under the precision-recall curve, approximated with a finite sum. 
%     \item[Normalized Discounted Cumulative Gain (NDCG)] is a measure that computes the weighted average of the relevance scores of the top retrieval results. The weights reduce the relevance value logarithmically, proportional to the position of the result \cite{distinguishability2013theoretical}. 
%     \item[Nearest Neighbor(NN)] \todo{add the description}.
% \end{description}

%  As the distance (relevance score) between the query and target we use an Euclidean distance in the embedded space, normalized to a unit hypersphere.

\subsection{Effect of sampling and rendering.} 
% In evaluation stage and retrieval, we extract the output of the penultimate layer of the encoder for each baseline, which is 512-dimensional, as the descriptor of each 3D object. 
% \clarify{Ling: I found only using classification loss to train can't get significant difference among sampling and rendering methods, so I train with CL+TCL. Suggest to move this evaluation to experiments part}

\begin{figure}[ht]
	\vspace{-2pt}
    \centering  
    % \fbox{\rule{0pt}{1.0in} \rule{0.9\linewidth}{0pt}}
    %\includegraphics[width=\linewidth]{image/render_sample.pdf}
    \includegraphics[width=\linewidth]{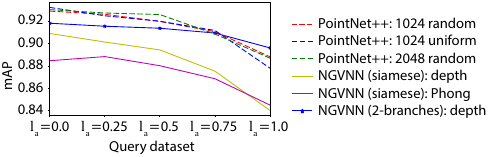}
    \caption{Comparison of sampling strategies and rendering styles, evaluated on synthetic datasets generated with different levels of abstractness $l_a$. All the baselines in this figure were trained using a combination: CL + TL.}
    \label{fig:representation}
    \vspace{-10pt}
\end{figure}

We compare different sampling methods for PointNet++ and rendering styles for NGVNN, described in Section \ref{sec:sampling_rendering}, when training with the classification loss and the triplet loss on the pairs of 3D shapes and 3D sketches rendered with $l_a$ set to 0.5. As shown in Figure \ref{fig:representation}, the uniform sampling for PointNet++ and depth rendering for NGVNN perform best, so we use these settings for the rest of the experiments. 

It can be seen from Figure \ref{fig:representation} that point-based representation (dashed lines) by far outperforms the multi-view representation (solid lines) when we use the siamese architecture for NGVNN. We thus for the rest of experiments use the heterogeneous architecture for NGVNN.

% \todo{Comment on observations on sampling, target as a curve network and different rendering strategies, as well as PointNet++ vs NVGNN baselines  Figure \ref{fig:representation}}

\subsection{Triplet loss vs triplet center loss}
\begin{figure}[ht]
	\vspace{-6pt}
    \centering  
    % \fbox{\rule{0pt}{1.0in} \rule{0.9\linewidth}{0pt}}
    %\includegraphics[width=\linewidth]{image/margin.pdf}
    \includegraphics[width=\linewidth]{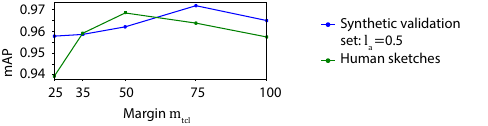}
    \caption{Evaluation of the performance of PointNet++, trained with CL+TCL, as a function of the different values of $m_{tcl}$, see Equation \ref{math:triplet_center}.}
    \label{fig:margin}
   	\vspace{-6pt}
\end{figure}

% \begin{figure}[ht]
%     \centering  
%     % \fbox{\rule{0pt}{1.0in} \rule{0.9\linewidth}{0pt}}
%     %\includegraphics[width=\linewidth]{image/margin.pdf}
%     \includegraphics[width=\linewidth]{image/margin.pdf}
%     \caption{Evaluation of the performance of PointNet++, trained with CL+TCL, as a function of the different values of $m_{tcl}$, see Equation \ref{math:triplet_center}. We choose $m_{tcl}=75$ based on valid set of synthetic sketches.}
%     \label{fig:margin}
% \end{figure}
%
The performance of the triplet (TL) and triplet center losses (TCL) is highly dependent on the selected values of the margin. For the triplet loss we select the margin $m_{tl}$ value as described in Section \ref{sec:metric_learning}. 
To select the optimal value of $m_{tcl}$, we train PointNet++ with CL and TCL on the synthetic dataset with $l_a = 0.5$ and different values of $m_{tcl}$ (Figure \ref{fig:margin}). 
Different values optimize the mAP on the synthetic, $m_{tcl} = 75$, and human, $m_{tcl} = 50$, test data.
% For the rest of the experiments in the paper we select the value $m_{tcl} = 75$ that optimizes the mAP on the synthetic validation set and performs reasonably on human data. Note, that the value $m_{tcl} = 50$ optimizes the mAP on human sketches. 
Figure \ref{fig:baselines} shows that the mAP of TL and TCL on our problem is comparable, while the Top-1 accuracy is much higher for the TL, showing that TL learns better inner-class variance and is more suitable for a fine-grained retrieval.

\begin{figure}[ht]
	\vspace{-6pt}
    \centering  
    % \fbox{\rule{0pt}{1.0in} \rule{0.9\linewidth}{0pt}}
    %\includegraphics[width=\linewidth]{image/margin.pdf}
    \includegraphics[width=\linewidth]{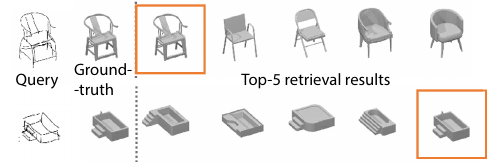}
    \caption{Example retrieval results with TL+Rec, the shapes matching the ground-truth are marked with orange boxes. Please see the supplemental for more examples.}
    \label{fig:retrieval}
    \vspace{-6pt}
\end{figure}

\begin{figure*}[t!]
    \centering  
    \includegraphics[width=\linewidth]{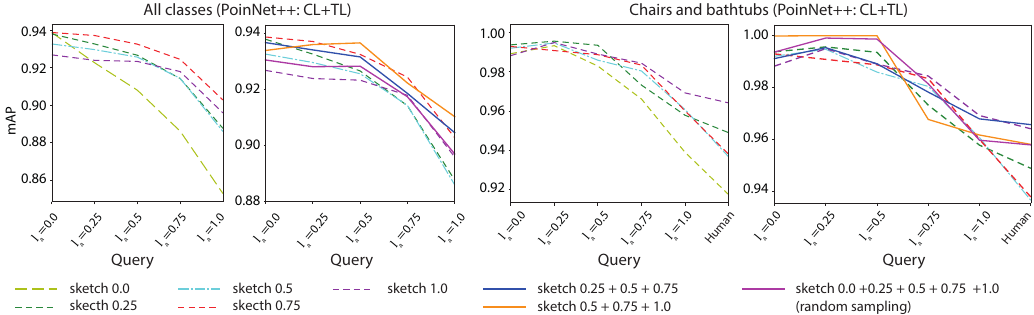}
    \caption{Comparison of the training datasets (Section \ref{sec:eval_abstraction}). Each line corresponds to a different training dataset. For the dataset which consists of all abstractness levels for each shape we randomly selected one of the abstractness levels.}
    \label{fig:abstraction}
\end{figure*}

\subsection{Regularization branch}
The reconstruction branch improves the mAp values for the TCL, while the mAP values for TL are similar with and without the reconstruction branch, both resulting in top performance. In Top-1 scores the regression branch significantly boosts the performance of the TL loss, resulting in the top performance on all synthetic and human datasets (Figure \ref{fig:baselines}). Consistent with the observation from the previous section, TCL optimizes the inter-class variance, which is further improved with the reconstruction loss; TL optimizes the intra-class variance and strongly benefits from our reconstruction branch (Table \ref{tab:my_label}). Some visual results of the retrieval are provided in Figure \ref{fig:retrieval}.

% When we train PointNet++ with $m_{tcl}$ values set to 75, the value that optimizes the validations set, then the combination of CL, TCL and Reconstruction loss gives the results that outperform all other baselines on all synthetic datasets, as shown in Figure  \ref{fig:baselines}, but slightly looses the performance of the network trained with CL and TCL only. When set $m_{tcl}$ values set to 50, the value that optimizes the performance of the network trained with the combination of the CL and TCL, an addition of the reconstruction loss allows to achieve the superior performance, boosting the mAP from $0.969$ to $0.983$.

% \paragraph{Numerical evaluation}

\subsection{Study of data abstraction level}
\label{sec:eval_abstraction}

In this section, we use different levels of abstraction to train the PointNet++ with the combination of the classification and triplet losses (Figure \ref{fig:abstraction}). It can be seen that training on the cleanest sketches ($l_a = 0.0$), yellow dashed lines, generalizes the worst to the other abstraction levels and human sketches, while training with the most abstract sketches ($l_a = 1.0$) results in a stable performance, though lower than the optimal for each of abstractness levels. It gives nearly optimal performance on human sketches. Interestingly, the performance on humans sketches when training on sketches with $l_a = 0.25$ outperforms the training on sketches with $l_a = 0.5$ and $l_a = 0.75$, what might indicate that human participants either did detailed and clean sketches or highly abstract ones. Drawing more certain conclusions would though require a larger set of human sketches, which includes more participants.

\begin{table}[]
    \resizebox{\linewidth}{!}{%
    \begin{tabular}{lcccc}
    \toprule  %
    Model& mAP & NDCG & NN &  Top-1\\
    \midrule  %
    CL+TL: 3D$\rightarrow$3D NGVNN (2-branch)& 0.934	& 0.966 & 0.934 & 0.018\\
    CL+TL: PointNet++& 0.937 & 0.968 & 0.946 & 0.258\\
    CL+TL+Rec: PointNet++& 0.961 &	0.979 &	0.964 & \textbf{0.329}\\
    CL+TCL: PointNet++ ($m_{tcl}$=75)& 0.964 & 0.979 & 0.964 & 0.060\\
    CL+TCL+Rec: PointNet++ ($m_{tcl}$=75)& 0.961 & 0.978 & 0.958 & 0.090 \\
    CL+TCL: PointNet++ ($m_{tcl}$=50)& 0.969 &	0.981 &	0.964 & 0.084\\
    CL+TCL+Rec: PointNet++ ($m_{tcl}$=50)& \textbf{0.983} &	\textbf{0.990} &	\textbf{0.982} &  0.138\\
    \bottomrule %
    \end{tabular}
    }
    \caption{Comparison of all the models on human sketches.}
    \label{tab:my_label}
    \vspace{-10pt}
\end{table}

% Best from previous section: Ours \\

% * 3D shape to 3D shape \\
% * Ours on clean curve networks \\
% * Ours on 0.25 \\
% * Ours on 0.5 \\
% * Ours on 0.75 \\
% * Ours on 0.25 + 0.5 + 0.75 \\
% * Ours on 0.0 + 0.25 + 0.5 + 0.75 + 1.0\

%\paragraph{Numerical evaluation}

\subsection{Ablation study: 2D sketch vs 3D sketch}
\label{sec:retrieval_2Dvs3D}

\begin{figure}[h!]
    \centering  
    \includegraphics[width=\linewidth]{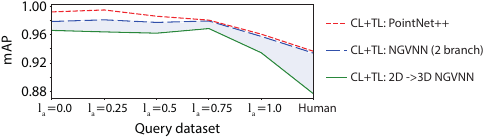}
    \caption{Comparison of the 2D sketch-based retrieval versus 3D sketch-based retrieval (Section \ref{sec:retrieval_2Dvs3D}). The baselines are trained on the dataset with $l_a = 0.5$.}
    \label{fig:2Dvs3D}
    \vspace{-10pt}
\end{figure}
We additionally compare the performance of the 3D sketch retrieval models to the performance of a 2D sketch retrieval model.
We use NGVNN as a backbone to build the 2D-sketch based shape retrieval baseline (2D$\rightarrow$3D NGVNN), where we use  NGVNN to encode 3D shape and additional branch to encode a 2D sketch, which is a common practice for 2D sketch/image based retrieval \cite{he2019view}. The 2D branch uses the same VGG11 backbone as NGVNN.
% and we change the output dimension of final linear layer to 512.
%by adding an independent branch for 2D sketch image, which is a common practice for 2D sketch/image based retrieval \cite{he2019view}. This baseline network is composed of one branch for 2D view and one branch for 3D shape. 
The input for the 2D sketch branch is the view rendered with the camera azimuth angle equal to $45^\circ$. It can be seen in Figure \ref{fig:2Dvs3D} that the 2D sketch-based retrieval is outperformed by all the models for 3D sketch-based retrieval, advocating for the 3D sketch-based retrieval.
%, as the one shown in Figure \ref{fig:rendering_views}.

%\subsubsection{Generalization of 3D to 3D shape models to 3D sketches}
%* Test how the models trained with 3D shape to 3D shape generalizes to 3D sketches.

% \subsection{Performance on human sketches}
% \todo{Test abstraction level on both synthetic and human sketches.}
% Human sketches Section \ref{sec:human_sketch}

% \subsection{Implementation details} 

% We implemented all models in Pytorch. Most rendering and deformation scripts are implemented in Blender. We conduct our experiments on a Linux workstation with one TeslaP40 GPU.

\section{Conclusion}
In this work we proposed the problem of 3D VR-sketch to 3D model retrieval. We first introduced a purpose-built VR environment to collect VR-sketches and conduct retrieval. We then collected a set of 3D human sketches for a subset of two shape classes from ModelNet10. We further proposed an approach for generating synthetic 3D sketches with different levels of abstraction, and demonstrated that the methods trained on our synthetic data generalize well to human sketches. Via a series of comprehensive evaluations, we find that compared to 3D shape-based retrieval, point-based shape representation is advantageous over the multi-view representation. At last, we propose an architecture with an additional sketch regularizing branch that leads to a superior performance over all the considered baselines, demonstrating the benefit of directly tackling the abstract nature of VR-sketches.
We hope to have offered a valid first stab at this new problem.

{\small
\bibliographystyle{ieee}
\bibliography{camera_ready}
}

\end{document}